\documentclass[conference]{IEEEtran}
\IEEEoverridecommandlockouts

\usepackage{booktabs} 
\usepackage{multirow} 
\usepackage{amssymb}  
\usepackage{makecell}
\usepackage{subcaption}
\usepackage{booktabs}
\usepackage{tabularx}
\usepackage{multirow}
\usepackage{graphicx}
\usepackage{xcolor}
\usepackage{cite}
\usepackage{amsmath,amssymb,amsfonts}
\usepackage{algorithmic}
\usepackage{graphicx}
\usepackage{textcomp}
\usepackage{xcolor}
\def\BibTeX{{\rm B\kern-.05em{\sc i\kern-.025em b}\kern-.08em
    T\kern-.1667em\lower.7ex\hbox{E}\kern-.125emX}}
\begin{document}

\title{DocIQ: A Benchmark Dataset and Feature Fusion Network for Document Image Quality Assessment\\

}

\author{
\IEEEauthorblockA{\textsuperscript{1}Zhichao Ma, \textsuperscript{1}Fan Huang, \textsuperscript{2}Lu Zhao, \textsuperscript{2}Fengjun Guo, \textsuperscript{1}Guangtao Zhai, \textsuperscript{1}Xiongkuo Min}
\IEEEauthorblockA{\textsuperscript{1}Shanghai Jiao Tong University, Shanghai, China}
\IEEEauthorblockA{\textsuperscript{2}INTSIG Information Co. Ltd, Shanghai, China}
\IEEEauthorblockA{\textsuperscript{1}\{august1, huangfan, zhaiguangtao, minxiongkuo\}@sjtu.edu.cn, \textsuperscript{2}\{lu\_zhao, fengjun\_guo\}@intsig.net}
}

\maketitle

\begin{abstract}
Document image quality assessment (DIQA) is an important component for various applications, including optical character recognition (OCR), document restoration, and the evaluation of document image processing systems. In this paper, we introduce a subjective DIQA dataset DIQA-5000. The DIQA-5000 dataset comprises 5,000 document images, generated by applying multiple document enhancement techniques to 500 real-world images with diverse distortions. Each enhanced image was rated by 15 subjects across three rating dimensions: overall quality, sharpness, and color fidelity. Furthermore, we propose a specialized no-reference DIQA model that exploits document layout features to maintain quality perception at reduced resolutions to lower computational cost. Recognizing that image quality is influenced by both low-level and high-level visual features, we designed a feature fusion module to extract and integrate multi-level features from document images. To generate multi-dimensional scores, our model employs independent quality heads for each dimension to predict score distributions, allowing it to learn distinct aspects of document image quality. Experimental results demonstrate that our method outperforms current state-of-the-art general-purpose IQA models on both DIQA-5000 and an additional document image dataset focused on OCR accuracy. 
\end{abstract}


\section{Introduction}
With the proliferation of digital technologies, document digitization has become indispensable for efficient storage, processing, and transmission of information. Despite the widespread use of paper documents, high-quality digital scans remain crucial for seamless integration between physical and digital formats. However, variations in capture devices and environmental conditions often degrade document image quality, necessitating robust enhancement techniques. Document Image Quality Assessment (DIQA) serves as a critical tool for evaluating these enhancement techniques, ensuring optimal readability and usability.

Existing research on Image Quality Assessment (IQA) has predominantly focused on natural scene images, as reflected in widely used datasets such as KonIQ-10k\cite{koniq10k}, CLIVE\cite{clive}, and CSIQ\cite{csiq}. However, these solutions prove inadequate for document images due to fundamental differences in structural and semantic characteristics. Document images exhibit distinct degradation patterns (e.g., blur, noise, uneven illumination) that demand specialized evaluation frameworks. While traditional DIQA methods often rely on handcrafted features such as gradient magnitudes\cite{b1} or directional sharpness metrics \cite{b2}, they often fail to generalize to complex, real-world distortions. In contrast, deep learning-based methods\cite{b3,b4} automatically learn multi-level features, demonstrating superior performance in general IQA tasks. Nevertheless, the lack of comprehensive document-specific datasets has hindered progress in data-driven DIQA.

To address these limitations, we introduce DIQA-5000, a novel document image quality assessment dataset comprising images with diverse distortion types. These images were processed using various document processing systems and subsequently evaluated through multi-dimensional human assessments, with final quality ratings annotated as mean opinion scores (MOSs). We further propose DocIQ, a DIQA model that integrates document layout features with multi-level image representations while effectively aggregating multi-rater scoring information. Experimental results demonstrate that our model achieves excellent performance on the proposed benchmark as well as another DIQA dataset..

\begin{figure*}[t]
    \centering
    \begin{subfigure}[b]{0.32\textwidth} 
        \includegraphics[width=\textwidth]{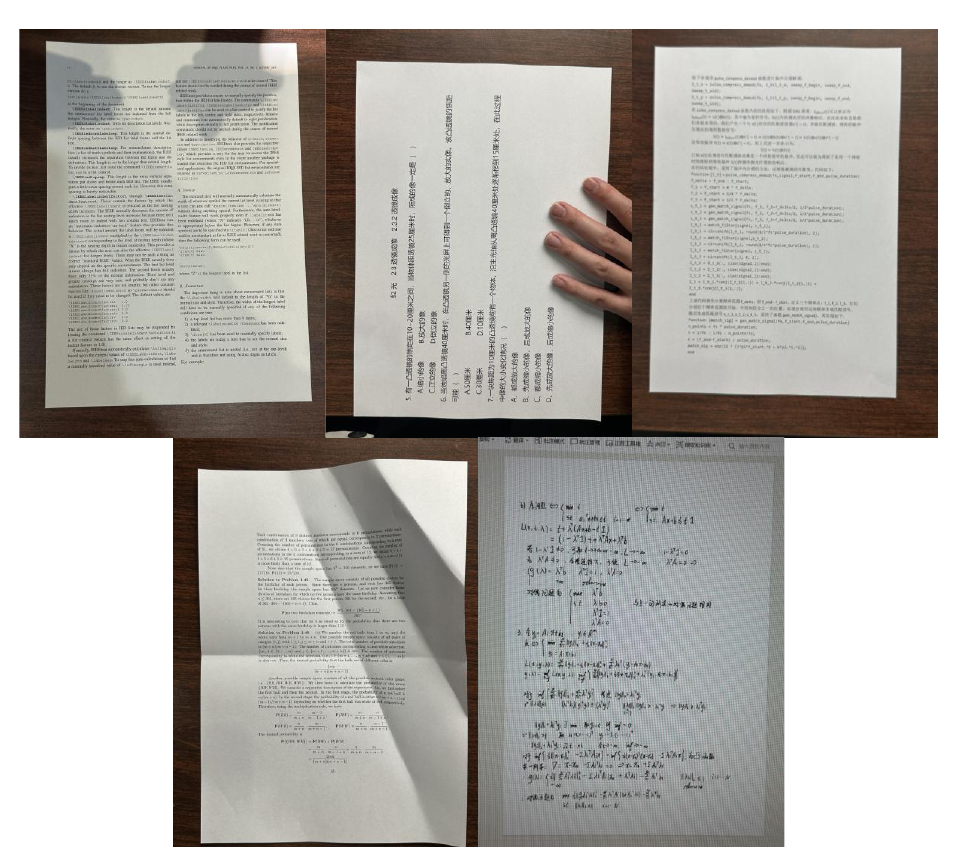}
        \caption{}
        \label{fig:subfig_a}
    \end{subfigure}
    \hfill 
    \begin{subfigure}[b]{0.32\textwidth}
        \includegraphics[width=\textwidth]{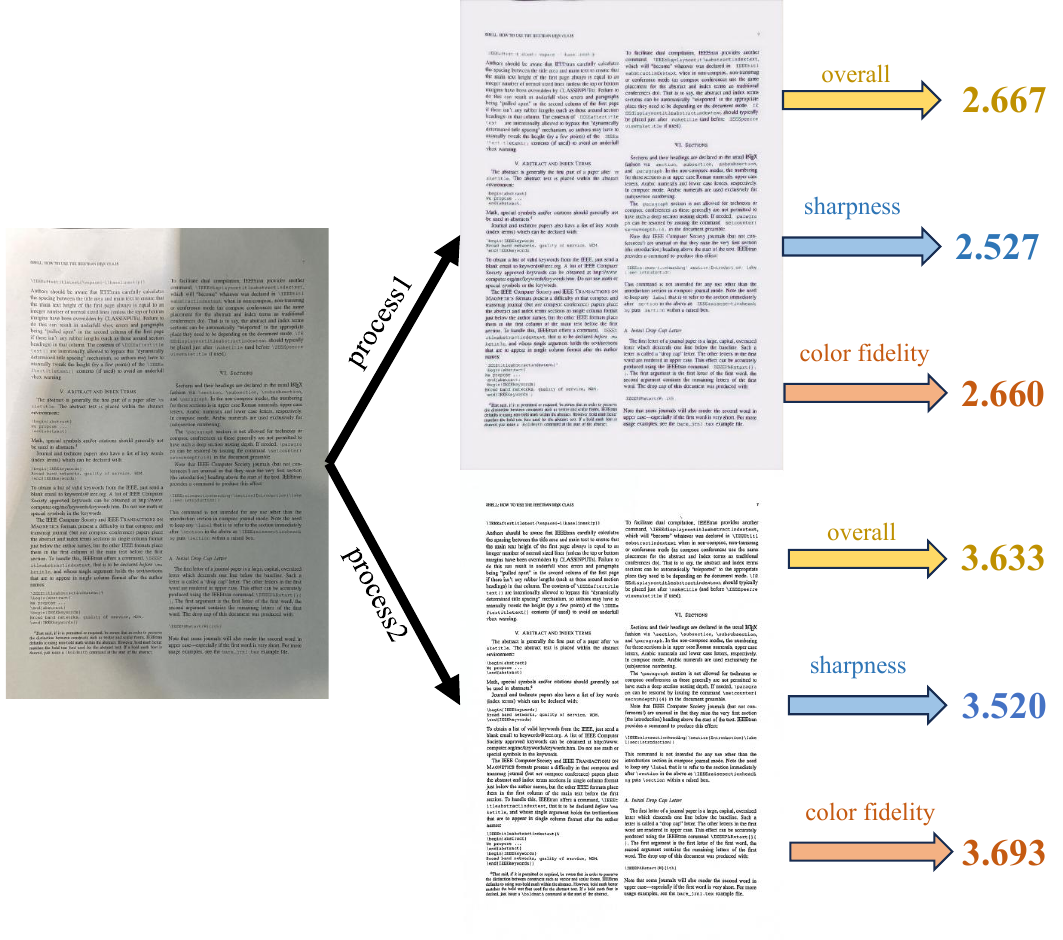}
        \caption{}
        \label{fig:subfig_b}
    \end{subfigure}
    \hfill
    \begin{subfigure}[b]{0.32\textwidth}
        \includegraphics[width=\textwidth]{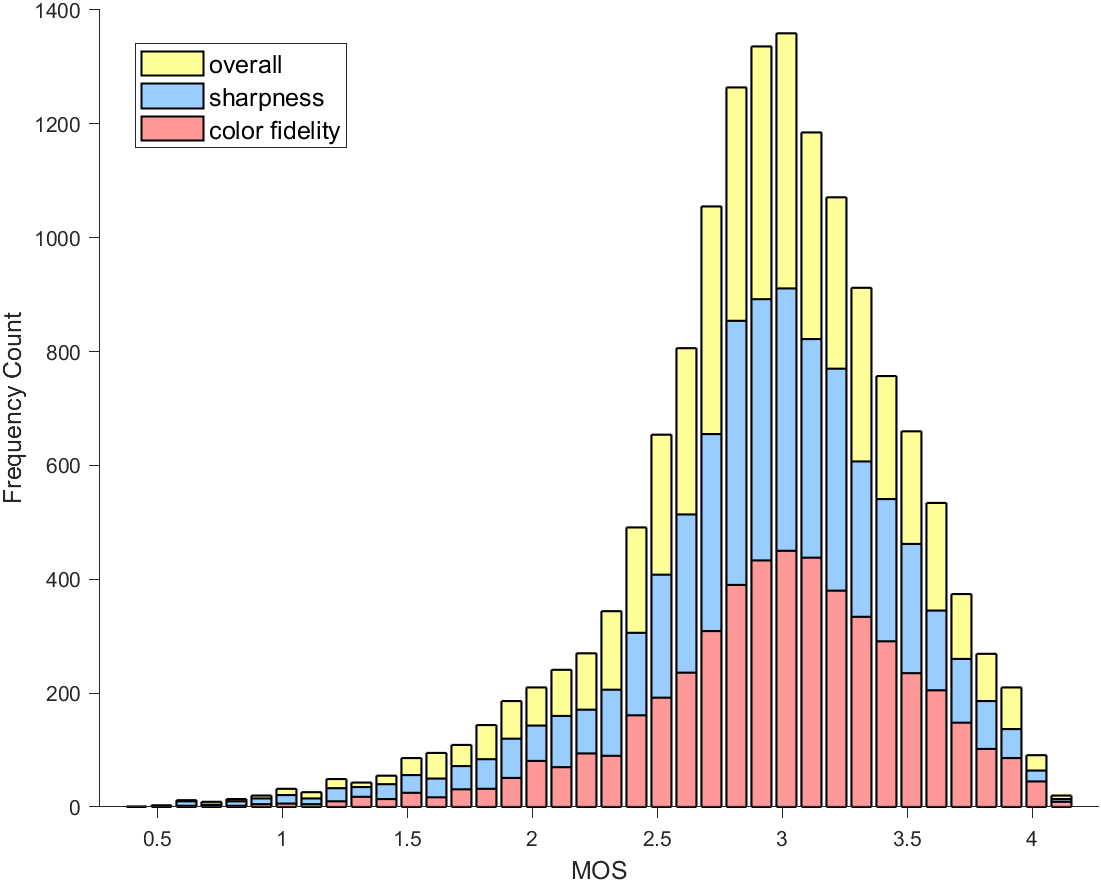}
        \caption{}
        \label{fig:subfig_c}
    \end{subfigure}
    \caption{Sample images and MOS distributions from the DIQA-5000 dataset. (a) shows examples of raw images for all 5 distortion types, arranged from left to right and top to bottom: shadow, occlusion, blurring, creases, and moiré patterns. (b) presents different enhanced versions of the same original image, generated through various processing pipelines. These images were rated by annotators across multiple quality dimensions. (c) shows the distributions of MOSs across the three evaluated dimensions for the entire dataset.}
    \label{fig:fig1}
\end{figure*}

\begin{figure}[t]
    \centering
    \includegraphics[width=0.95\columnwidth]{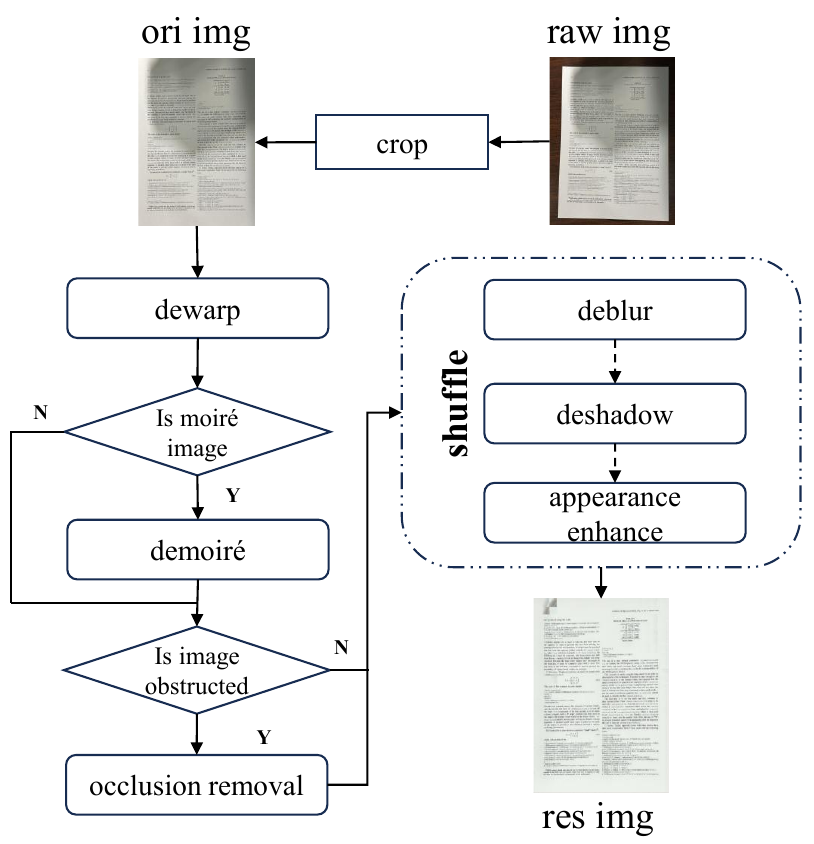}
    \caption{Document image processing pipeline. Each stage includes multiple available methods—dewarp (3 options), demoiré (2), occlusion removal (2), deblur (3), deshadow (4), and appearance enhancement (9)—and different processing flows are generated through random combinations.}
    \label{fig:fig2}
\end{figure}

\begin{figure*}[t] 
    \centering
    \includegraphics[width=0.95\textwidth]{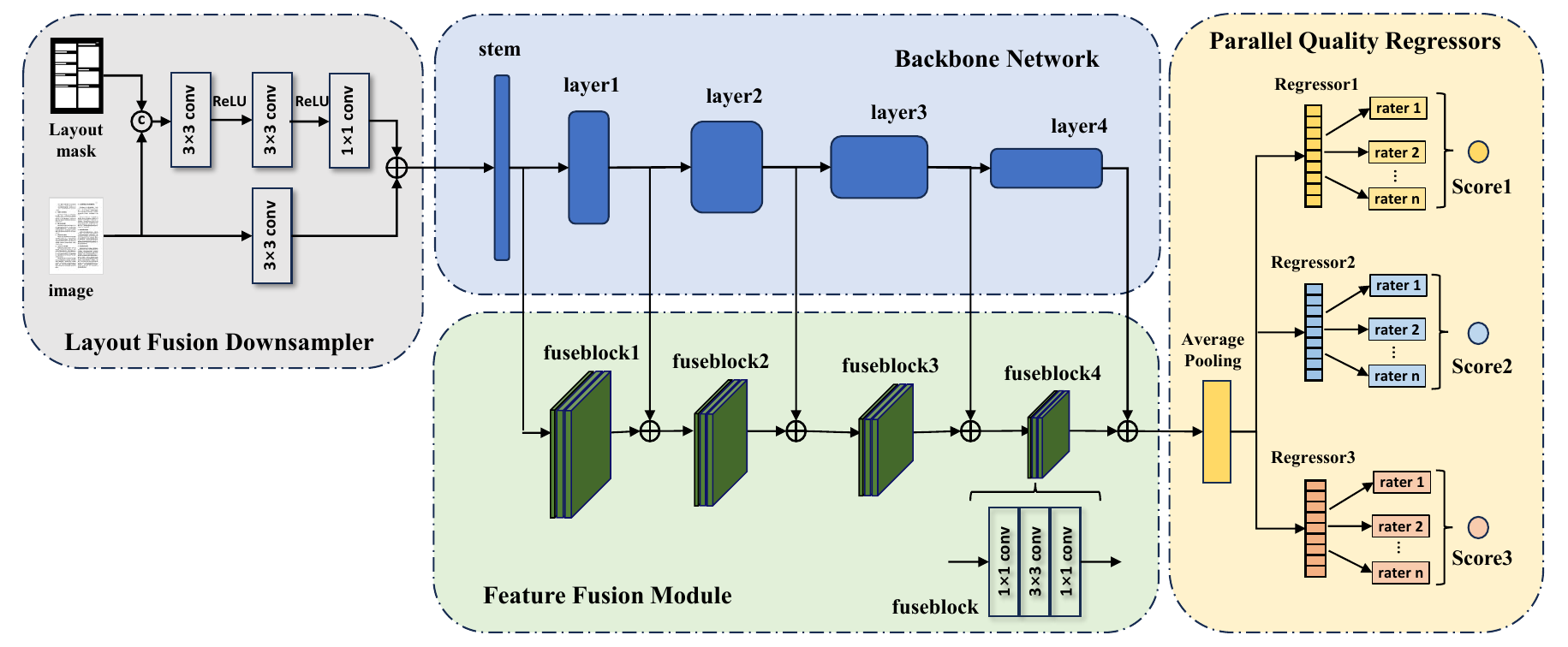}
    \caption{The network architecture of the proposed DocIQA model, which consists of 4 key components.}
    \label{fig:fig3}
\end{figure*}

\section{DIQA-5000}\label{s1}
In this section, we introduce DIQA-5000, a novel subjective dataset designed to assess the performance of document processing systems across diverse degradation scenarios.

\subsection{Document Image Acquisition}
We curated a representative set of document images from publicly accessible PDFs, spanning textual, tabular, and mixed-content layouts to ensure diversity. These documents were printed at 300 dpi to create original paper documents.

To simulate real-world capture conditions, we introduced five distortion types:

\begin{enumerate}
    \item Shadow: uneven lighting during smartphone capture;
    \item Occlusion: partial obstruction by objects;
    \item Blurring: motion blur or defocus effects;
    \item Creases: physical folds on printed documents;
    \item Moiré Patterns: artifacts from capturing screens displaying documents.
\end{enumerate}

For each distortion category, 100 images were captured using a specified mobile phone, yielding 500 distorted images in total. Example distortions are illustrated in Fig. 1(a). 

\subsection{Document Image Processing Pipeline}
To comprehensively evaluate document enhancement methods, we developed an image processing pipeline that applies six core operations in randomized combinations: dewarp, demoiré, occlusion removal, deblur, deshadow, and appearance enhancement. Dewarp aims to correct distortions. Demoiré and occlusion removal are designed to eliminate moiré patterns and obstructions, respectively. Deblur targets motion or defocus blur to produce sharper, more readable document images. Deshadow removes shadows caused by lighting conditions or occlusions. Appearance enhancement aims to enhance the visual quality to resemble that of scanned or digitally generated PDF. The pipeline incorporates both open-source implementations\cite{b11,b12,b13,b14,b15,b16,b17,b18} and commercial SDKs, ensuring diversity in processing methodologies.

The document processing pipeline is illustrated in Fig. \ref{fig:fig2}. The workflow begins with document boundary detection and background removal. Then, the original image goes through a multi-step processing flow. At each processing stage, the system randomly selects among available algorithms or skips the operation, with the execution order of deblurring, deshadowing and enhancement stages being randomized. 
This stochastic approach generates 10 distinct enhanced versions per input image, producing a total of 5,000 enhanced document images from the original 500 captures. The randomized design ensures comprehensive coverage of enhancement scenarios while preventing algorithmic bias in the dataset construction.

\subsection{Subjective Quality Assessment}
The subjective evaluation protocol includes three rating dimensions: overall quality, sharpness, and color fidelity.
Twenty-three experienced subjects participated in the evaluation. The 5,000 images were divided into five balanced batches, with 15 raters assigned to each batch. As a result, each image received 15 independent scores per rating dimension. We also implemented data cleaning to remove inconsistent or unreliable ratings according to ITU-R BT.500\cite{series2012methodology} for each batch. Fig. 1(b) presents representative annotated samples from our dataset, showing different enhanced versions of the same original document image along with their corresponding multi-dimensional quality ratings. Fig. 1(c) displays the MOS distributions for all three rating dimensions (overall quality, sharpness, and color fidelity), revealing the dataset's coverage of quality variations.

\section{DocIQ Model}
The proposed DocIQ architecture, illustrated in Fig.~\ref{fig:fig3}, comprises four key components: layout fusion downsampler, backbone network, feature fusion module, and parallel quality regressors.

\subsection{Layout Fusion Downsampler}
To address the computational challenges of processing high-resolution document images, we developed a lightweight downsampling module that incorporates document layout semantics. The module employs a dual-path architecture where the primary path performs conventional spatial downsampling while the secondary path processes a concatenated input of the original image and its semantic layout mask (identifying text regions, tables, and figures).  This design reduces computational complexity  while enhancing feature relevance through semantic region focusing and preserving crucial spatial relationships. The layout masks are generated using pretrained document layout detection model \cite{layout1,layout2}.

\subsection{Backbone Network and Feature Fusion Module}
While conventional CNNs like VGG\cite{b9} and ResNet\cite{b10} excel at hierarchical feature extraction, their progressive downsampling inherently discards low-level details essential for quality assessment\cite{gao2017deepsim}. To better align with human visual perception, we introduce the feature fusion module in our DocIQ model. This module progressively fuses multi-scale features extracted from different stages of the backbone. Specifically, low-level spatial features are combined with high-level semantic features through a series of lightweight hyper-structures. Each hyper-structure consists of bottleneck convolutions that compress the channel dimension, apply spatial transformations, and then restore the output dimension.

The fusion process starts from the lowest feature map and proceeds layer-by-layer. At each stage, the fused representation is added to the corresponding high-level feature from the backbone. The final output is a compact yet semantically enriched global feature that captures both structural fidelity and semantic relevance. This fused representation is then passed to a set of parallel regression heads to predict multiple quality scores, enabling fine-grained document image quality assessment.

\subsection{Parallel Quality Regressors}

In DIQA, it is common to evaluate each image across multiple quality dimensions with multiple raters. To better capture the quality distribution of these multi-dimensional features, we adopt a multi-heads regression architecture. The framework utilizes independent regression heads for each quality dimension, predicting individual rater scores and aggregating them into a final MOS.

Specifically, the global feature obtained from the feature fusion module is fed into a set of parallel fully connected regressors. Each regression head comprises two linear layers: a shared first layer and dimension-specific second layers that separately predict scores for each rater. These layers map high-level features to scalar scores for their respective quality dimensions, and output predicted scores for each rater. This design enables the model to learn specialized representations for different degradation types while benefiting from shared backbone features. Moreover, the architecture maintains robust performance even when partial rater information is unavailable, as it learns the complete quality distribution.

\begin{table*}[t]
    \centering
    \caption{Performance comparison on DIQA datasets. Best and second-best scores are shown in bold and underlined, respectively. Missing values are denoted by `-' due to the absence of score distribution in the SmartDoc-QA dataset.}
    \label{tab:results}
    \small
    \setlength{\tabcolsep}{4pt} 
    \renewcommand{\arraystretch}{1.2} 
    \begin{tabularx}{\textwidth}{l *{10}{>{\centering\arraybackslash}X}}
        \toprule
        & \multicolumn{6}{c}{DIQA-5000} & \multicolumn{4}{c}{SOC\cite{b19}} \\
        \cmidrule(r){2-7} \cmidrule(r){8-11}
        Method & \multicolumn{2}{c}{Overall} & \multicolumn{2}{c}{Sharpness} & \multicolumn{2}{c}{Color Fidelity} & \multicolumn{2}{c}{CACC} & \multicolumn{2}{c}{WACC} \\
        \cmidrule(r){2-3} \cmidrule(r){4-5} \cmidrule(r){6-7} \cmidrule(r){8-9} \cmidrule(r){10-11}
        & PLCC & SRCC & PLCC & SRCC & PLCC & SRCC & PLCC & SRCC & PLCC & SRCC \\
        \midrule
        DBCNN\cite{b3} & 0.5869 & 0.5421 & 0.6163 & 0.6037 & 0.6335 & 0.6399 & 0.8816 & 0.8780 & 0.8899 & 0.8722 \\
        HyperIQA\cite{b4} & 0.8437 & 0.8024 & 0.8542 & 0.8197 & 0.8439 & 0.8155 & 0.8900 & 0.8828 & 0.8919 & 0.8557 \\
        MUSIQ\cite{b5} & 0.8585 & \underline{0.8554} & 0.8698 & \underline{0.8460} & 0.8460 & 0.8383 & 0.8783 & 0.8651 & 0.8813 & 0.8653 \\
        RichIQA\cite{b6} & \underline{0.8660} & 0.8541 & 0.8770 & 0.8357 & 0.8622 & \underline{0.8557} & -- & -- & -- & -- \\
        StairIQA\cite{b7} & 0.8502 & 0.8004 & 0.8671 & 0.8359 & \underline{0.8691} & 0.8476 & \underline{0.9138} & \underline{0.8921} & \underline{0.8980} & \underline{0.8857} \\
        TReS\cite{b8} & 0.8628 & 0.8080 & \underline{0.8800} & 0.8267 & 0.8658 & 0.8338 & 0.8893 & 0.8841 & 0.8753 & 0.8701 \\
        \midrule 
        DocIQ & \textbf{0.9083} & \textbf{0.8832} & \textbf{0.9006} & \textbf{0.8615} & \textbf{0.8907} & \textbf{0.8666} & \textbf{0.9218} & \textbf{0.9086} & \textbf{0.9107} & \textbf{0.8989} \\
        \bottomrule
    \end{tabularx}
\end{table*}

\begin{table}[htbp]
\centering
\small 
\setlength{\tabcolsep}{5pt} 
\caption{Ablation study for different modules across different dimensions on the DIQA-5000 dataset.}
\label{tab:ablation}
\begin{tabular}{ccc|c cc c}
\toprule
\thead{Layout Fusion\\ Downsampler} & \thead{Feature\\ Fusion} & \thead{Multi-Raters\\ Strategy} & \thead{Overall} & \thead{Sharpness} & \thead{Color\\ Fidelity} \\
\midrule
$\checkmark$ & $\checkmark$ & $\checkmark$ & 0.8832 & 0.8615 & 0.8666 \\
$\checkmark$ & $\checkmark$ & $\times$     & 0.8636 & 0.8545 & 0.8553 \\
$\times$     & $\checkmark$ & $\checkmark$ & 0.8696 & 0.8481 & 0.8589 \\
$\checkmark$ & $\times$     & $\checkmark$ & 0.8448 & 0.8293 & 0.8401 \\
$\times$     & $\times$     & $\checkmark$ & 0.8162 & 0.7901 & 0.8137 \\
\bottomrule
\end{tabular}
\end{table}

\section{Experiments}

\subsection{Experimental Settings}
The evaluation utilizes two DIQA benchmarks. The DIQA-5000 dataset, proposed in this work, contains 5,000 document images with comprehensive multi-dimensional quality annotations. We also include the public SmartDoc-QA dataset\cite{b19}, which comprises 2,130 smartphone-captured document images (after cleaning), with quality assessed through OCR performance metrics: Character ACCuracy (CACC) and Word ACCuracy (WACC).

The proposed approach is benchmarked against six representative no-reference IQA methods. All comparison methods were retrained and tested using their original configurations to ensure fair evaluation. We employ two well-established correlation metrics to quantitatively assess model performance: the Spearman Rank Correlation Coefficient (SRCC) measures monotonic relationships between predicted and ground truth scores, while the Pearson Linear Correlation Coefficient (PLCC) evaluates linear correlations.

The experimental implementation employs an 80\%-20\% training-testing split for both datasets. Our model builds upon a ResNet50 backbone pretrained on ImageNet, processing 1600×1600 resolution images augmented with layout masks. The training protocol uses Adam optimization with an initial learning rate of $2 \times 10^{-4}$ and step decay scheduling (step size: $10$ epochs, decay factor: $0.6$), executed over 60 epochs with batch size 20 on NVIDIA A10 GPUs.

\subsection{Comparative Results}
The experimental results in Table~\ref{tab:results} demonstrate significant advantages of our proposed method over existing approaches across both evaluation datasets. On the DIQA-5000 benchmark, our model achieves an average SRCC of 0.8704 and PLCC of 0.8999 across all rating dimensions, outperforming the compared IQA models. Moreover, the compared models adopt a single-output-head architecture, which can only predict one specific quality score at a time. In contrast, our model incorporates independent multi-quality heads, enabling it to produce scores across multiple dimensions from a single input image, which is more efficient and flexible for multi-dimensional quality assessment. On the SmartDoc-QA dataset, the method maintains strong correlation with OCR-based metrics (CACC SRCC=0.9086, WACC SRCC=0.8989). Despite the absence of multi-rater subjective annotations in OCR-based datasets, our model effectively aligns perceptual quality assessment with practical readability, demonstrating the effectiveness of layout modeling and feature fusion.

\subsection{Ablation Study}
Ablation studies were conducted to evaluate the contributions of key architectural components, with results summarized in Table~\ref{tab:ablation}. Removing the layout fusion downsampler led to an average SRCC decrease of 0.0115 across all dimensions, highlighting the importance of structural awareness. The notable improvement in sharpness estimation may stem from the model’s ability to capture clarity in semantically important regions like text, requiring joint consideration of semantics and edges. Disabling feature fusion caused an average 0.0323 SRCC drop, confirming the benefit of integrating low-level and high-level features. Replacing the multi-rater strategy with direct MOS regression reduced SRCC by 0.0126 on average, demonstrating that modeling rater variance enhances prediction robustness. These results validate the effectiveness of the proposed architectural design.

\section{Conclusion}
In this work, we make two key contributions to document image quality assessment (DIQA). First, we introduce DIQA-5000, a multi-dimensional subjective dataset with fine-grained quality ratings across three dimensions for 5,000 document images, covering diverse distortions and enhancement techniques. Second, we introduce DocIQ, a novel DIQA model that leverages multi-level and layout-aware features through dedicated regressors for each quality dimension, outperforming existing general purpose IQA methods.

\bibliographystyle{IEEEtran}
\bibliography{references}{}

\end{document}